\documentclass[10pt,twocolumn,letterpaper]{article}

\usepackage[pagenumbers]{cvpr} 

\usepackage[dvipsnames]{xcolor}

\definecolor{cvprblue}{rgb}{0.21,0.49,0.74}
\usepackage[pagebackref,breaklinks,colorlinks,citecolor=cvprblue]{hyperref}

\usepackage{amsmath} 
\usepackage{amsfonts} 
\usepackage{amssymb} 
\usepackage{bm}
\usepackage{multirow}
\usepackage{makecell}
\usepackage{graphicx}
\usepackage{booktabs}

\usepackage{xcolor}

\def\paperID{5923} 
\def\confName{CVPR}
\def\confYear{2025}

\title{Fleximo: Towards Flexible Text-to-Human Motion Video Generation}

\author{Yuhang Zhang\textsuperscript{\rm 1, 2}, Yuan Zhou\textsuperscript{\rm 2}, Zeyu Liu\textsuperscript{\rm 2, 3}, Yuxuan Cai\textsuperscript{\rm 2}, Qiuyue Wang\textsuperscript{\rm 2}, Aidong Men\textsuperscript{\rm 1}, Huan Yang\textsuperscript{\rm 2*}\\
\textsuperscript{\rm 1} Beijing University of Posts and Telecommunications, 
\textsuperscript{\rm 2} 01.AI,
\textsuperscript{\rm 3} Tsinghua University
}

\begin{document}

\twocolumn[{%
\renewcommand\twocolumn[1][]{#1}%
\maketitle
\begin{center}
\includegraphics[width=0.95\linewidth]{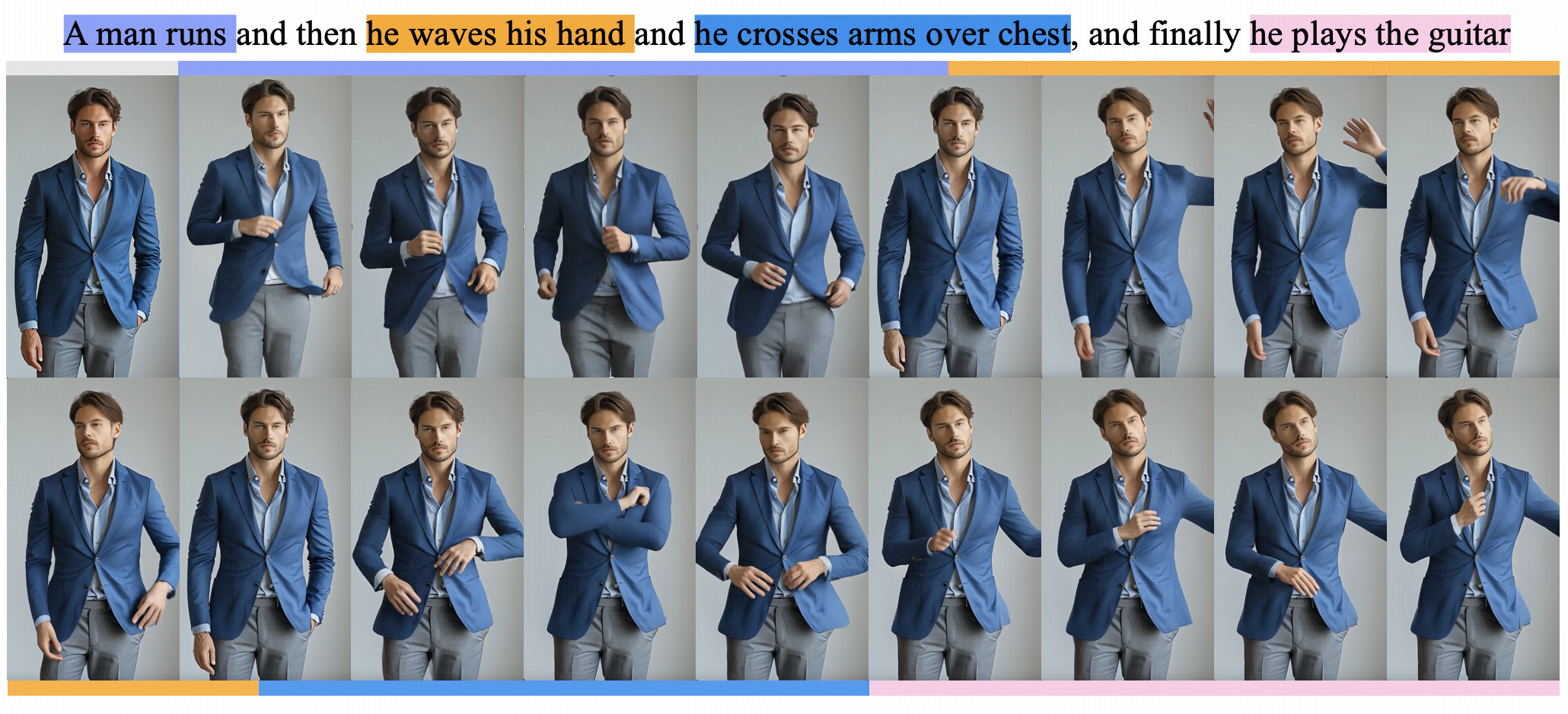}
\vspace{-3mm}
\captionof{figure}{Given a reference image and a motion text, our method \textbf{Fleximo} can generate motion videos containing the reference identity performing the motions described in the motion text. The reference image is shown at the first frame, the input motion text is on the top of the figure. Different colors mark the slices of different motion segments.}
\label{teaser}
\end{center}
}]

\footnotetext{*Corresponding author (hyang@fastmail.com)}

\begin{abstract}
\vspace{-2mm}
\noindent{Current methods for generating human motion videos rely on extracting pose sequences from reference videos, which restricts flexibility and control. Additionally, due to the limitations of pose detection techniques, the extracted pose sequences can sometimes be inaccurate, leading to low-quality video outputs.
We introduce a novel task aimed at generating human motion videos solely from reference images and natural language. This approach offers greater flexibility and ease of use, as text is more accessible than the desired guidance videos. However, training an end-to-end model for this task requires millions of high-quality text and human motion video pairs, which are challenging to obtain. To address this, we propose a new framework called \textbf{Fleximo}, which leverages large-scale pre-trained text-to-3D motion models. This approach is not straightforward, as the text-generated skeletons may not consistently match the scale of the reference image and may lack detailed information. To overcome these challenges, we introduce an anchor point based rescale method and design a skeleton adapter to fill in missing details and bridge the gap between text-to-motion and motion-to-video generation. We also propose a video refinement process to further enhance video quality. A large language model (LLM) is employed to decompose natural language into discrete motion sequences, enabling the generation of motion videos of any desired length.
To assess the performance of Fleximo, we introduce a new benchmark called MotionBench, which includes 400 videos across 20 identities and 20 motions. We also propose a new metric, MotionScore, to evaluate the accuracy of motion following. Both qualitative and quantitative results demonstrate that our method outperforms existing text-conditioned image-to-video generation methods. All code and model weights will be made publicly available.}
\vspace{-6mm}
\end{abstract}    
\section{Introduction}
\label{sec:intro}

Human motion video generation has become a prominent and rapidly evolving field within computer vision~\cite{wang2024disco, chang2023magicpose, xu2024magicanimate, hu2024animate, zhu2024champ, zhang2024mimicmotion}, with a primary focus on generating human videos driven by reference images and pose videos. Traditional methods utilize pose sequences extracted from human pose videos to guide motion in generated outputs. Although these approaches have yielded promising results, they impose substantial limitations regarding flexibility and precision. Relying on pose sequences restricts user control, as obtaining videos with specific desired motions can be both time-consuming and restrictive. Moreover, the accuracy of pose extraction techniques is often compromised by various factors such as occlusions or low-resolution inputs~\cite{zhang2024mimicmotion, yang2023effective}, resulting in corrupted or imprecise pose sequences that degrade the quality of the generated human motion videos.

To introduce a more flexible and user-friendly approach to human motion video generation, we propose a new task, \textbf{text-to-human motion video generation}, that requires only a reference image and natural language as inputs. This task significantly enhances user control by allowing motion to be directed through text, eliminating the need to source pre-existing videos with specific motion patterns. However, this new task also introduces unique challenges. Training a model to generate human motion videos based solely on text and reference images, i.e. text-conditioned image-to-video, requires an extensive number of text-video pairs, potentially in the millions or even billions~\cite{zhou2024allegro, polyak2024movie, liu2024sora}, making the process resource-intensive. Especially, there are currently no public datasets dedicated to high-quality, captioned human-motion-centric videos of this scale. Thus existing text-conditioned image-to-video approaches continue to face challenges such as limited motion generation with small movements and identity inconsistency with the reference image~\cite{zhang2023i2vgen, chen2023videocrafter1, xing2023dynamicrafter}.

Recent text-to-motion approaches~\cite{zhang2022motiondiffuse, jiang2023motiongpt, wang2024motiongpt} have achieved impressive results in generating 3D motion from text. In contrast to text-video data, text-3D mesh data benefits from established datasets~\cite{zhang2023generating, guo2022generating, chen2023executing}. Building on this, we introduce \textbf{Fleximo}, \textbf{a text-to-human motion video generation framework} that utilizes pretrained text-to-3D models by projecting 3D meshes onto 2D skeletons to guide video generation. However, achieving flexible and high-quality generation requires overcoming two main challenges: (1) the fixed scale of the generated 2D skeleton, which may not match the reference image, and (2) the limited detail in the 2D skeleton, particularly in areas like hands and face.

To address the first problem, we propose an \textbf{anchor point based rescale} method that uses keypoints from the reference image and frame-by-frame affine transformations for scale adjustment. Since affine transformations cannot fully match body part lengths, we further designate the neck as the anchor and individually adjust each connected keypoint, improving alignment with the reference image.
To solve the second problem, we develop a \textbf{skeleton adapter} that adds realistic detailed movements to sparse skeleton videos, creating a complete skeleton video with accurate hand motions.

We generate human motion videos with a motion-to-video model that utilizes the detailed 2D skeleton videos for motion guidance and a reference image for identity. This process can be iterative, with generated videos re-entered into the model for further refinement. To achieve complex motion sequences generation, we incorporate a large language model (LLM), LLaMA-7B~\cite{touvron2023llama}, to process long text inputs by segmenting them into discrete motion parts, enabling our framework to generate motion sequences and videos of any desired length.

Given the limited prior work on text-to-human motion video generation, we introduce \textbf{MotionBench}, a benchmark with 400 videos across 20 identities and 20 motions. We evaluate methods on identity fidelity, video quality, and introduce a new metric \textbf{MotionScore} for assessing motion-text alignments. 

The main contributions of our paper are as follows:

\begin{figure*}[t]
  \centering
  \includegraphics[width=1.\linewidth]{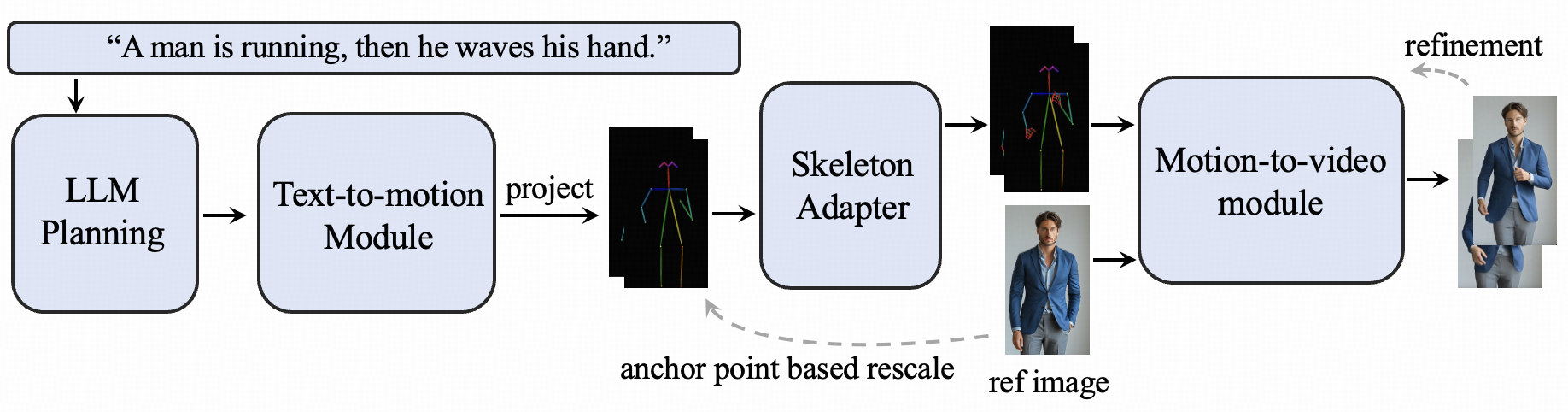}
    \caption{The framework of Fleximo.  
We use LLM to plan long motion texts. The text-to-motion module generates 3D mesh vertices corresponding to the motion texts. Then, these vertices are projected into 2D space. The 2D skeleton points are scaled based on the anchor point and formulated as skeleton videos. These skeleton videos are input into the skeleton adapter for detail completion. The output skeleton video and the reference image are used as guidance for human motion video generation. A video refinement process can further improve the generated video quality.
}
\vspace{-3mm}
    \label{pipeline}
\end{figure*}

\begin{itemize}
\item We introduce \textbf{a new task for generating high-quality human motion videos from text and reference images}, offering a more flexible and user-friendly approach by allowing direct control via natural language.
\item We propose a novel text-to-human motion video generation framework, \textbf{Fleximo}, that avoids the need for large text-video paired data. The framework leverages the large-data pretrained text-to-3D motion model and bridges the gap between text-to-motion and motion-to-video generation model.
\item We propose a benchmark, \textbf{MotionBench}, with various human identities and motions to evaluate motion video generation methods, along with a new metric, \textbf{MotionScore}, to assess motion-text alignments.
\end{itemize}

\section{Related Work}
\label{sec:formatting}
\subsection{Text-to-Motion Generation} 
Text-to-motion generation translates human language into physical actions, represented as skeletal animations, motion capture data, or motion descriptors~\cite{Guo_2022_CVPR, tevet2022motionclip, petrovich2022temos, zhang2022motiondiffuse, liu2024motionrl}. Its intuitive, user-friendly design makes it especially valuable. MDM~\cite{tevet2023human} introduces a diffusion-based generative model~\cite{ho2020denoising}, trained separately on various motion tasks, while MLD~\cite{chen2023executing} extends latent diffusion models~\cite{song2020denoising, rombach2022high} to generate motion from different conditional inputs. T2M-GPT~\cite{zhang2023generating} explores a generative framework combining VQ-VAE~\cite{van2017neural} and Generative Pre-trained Transformers (GPT)~\cite{radford2018improving} for motion generation.
Although these methods successfully generate 3D motion meshes from text, they are unsuitable for video generation, as they operate in 3D space, which leads to the scale misalignment and faces and hands information missing. In this paper, we propose the anchor point based rescale to address the scale misalignment. We fine-tune video generation model and propose a skeleton adapter to compensate the missing face and hand information, which bridges the gap between text-to-motion and video generation.

\subsection{Motion-to-Video Generation} 
Motion-to-video generation involves creating coherent video sequences based on input motion information, which can come from various sources such as motion capture data or skeletal animations. Disco~\cite{wang2024disco} introduces a Pose and Background ControlNet to control character movements and maintain background consistency. MagicPose~\cite{chang2023magicpose} enhances generation quality using an Appearance Control Model. Similarly, MagicAnimate~\cite{xu2024magicanimate} replaces OpenPose with DensePose, following a comparable approach. AnimateAnyone~\cite{hu2024animate} adopts a two-stage training process, with separately trains the ReferenceNet and motion layers, ensuring better character ID consistency. Champ~\cite{zhu2024champ} enriches video generation by incorporating depth images, normal maps, and semantic maps from SMPL sequences, alongside skeleton-based motion guidance. MimicMotion~\cite{zhang2024mimicmotion} proposes confidence-aware pose guidance and regional loss amplification to handle noise from pose detectors, achieving high-quality human dancing videos. While these methods produce high-quality motion videos, they rely on acquiring skeleton or mesh sequences for conditioning, which can be less user-friendly and controllable. In this paper, we aim to simplify the process by generating high-quality human motion videos using motion texts instead of skeleton or mesh sequences.

\section{Method}

\subsection{Overview}
Our goal is to animate a reference human image to follow motions described in text. We first use an LLM (LLaMA-7B~\cite{touvron2023llama}) to parse the input text, which may include multiple motions, breaking it into sequential motion segments. A text-to-motion module (T2M-GPT~\cite{zhang2023generating}) then generates 3D mesh vertices for each motion segment, projected into 2D space. After applying an affine transformation~\cite{jahne2005digital}, we use an anchor point to rescale body length to match the reference image, creating skeleton video sequences. These are processed by a skeleton adaptor for motion-to-video generation. The adapted skeleton videos, combined with the reference image, generate an initial anchor video, guiding the final refined motion video generation. Fig.~\ref{pipeline} illustrates the framework of Fleximo.

\subsection{Anchor Point Based Rescale}
The simplest baseline for solving our task is combining text-to-motion (T2M) model and motion-to-video (M2V) model. The T2M model like T2M-GPT can generate 3D meshes according to text. Given the input motion text $P$, the T2M model $F(P; \theta)$ outputs sets of vertices $\{ V^{3D}_{i} \}_{i=1}^{K}$, $V^{3D}_{i} \in \mathbb{R}^3$, which form meshes representing the human motion, as expressed below:

\begin{equation}
F(P; \theta) = \{ V^{3D}_{1}, \dots, V^{3D}_{K} \}
\end{equation} $\theta$ represents the parameters of the T2M model, and $K$ is the number of vertices in the meshes. Then the 3D meshes can be projected into 2D skeleton following 
\begin{equation}
\mathbf{V}^{2D}_{i} = 
\begin{pmatrix}
1 & 0 & 0 \\
0 & 1 & 0 \\
\end{pmatrix} 
\mathbf{V}^{3D}_{i}
=
\begin{pmatrix}
x_{i} \\
y_{i} \\
\end{pmatrix}
\end{equation}
$(x_{i}, y_{i})$ is the keypoint of human. We then plot each frame of keypoints into a image and transform all frames into a skeleton video. The M2V model like MimicMotion~\cite{zhang2024mimicmotion} takes the 2D skeleton video for motion guidance and a reference image for ID guidance and generates the human motion video.

However, directly combining these two models cannot generate human motion videos with high-quality as there are mainly two problems: the scale of the skeleton video might not be the same as the reference image and the skeleton video does not contain face or hands information. To solve these problems, we propose several techniques. For the scale problem, we propose to first use affine transformation to roughly align the generated skeleton keypoints and the keypoints detected from the reference image. Specifically, let $y^{d}$ and $y^{r}$ represent the y-coordinates of detected bodies and the reference body. The scaling factor \(a_y\) and offset \(b_y\) are computed using a linear fit:

\begin{equation}
a_y, b_y = \text{polyfit}(y^{d}, y^{r}, 1)
\end{equation}
Then, we adjust \(a_y\) by the frame aspect ratio and the \text{height} and \text{width} of the reference image to compute the x-coordinate scaling factor \(a_x\), :
\begin{equation}
a_x = \frac{a_y}{\frac{f_h}{f_w} \times \frac{\text{height}}{\text{width}}}
\end{equation}
where \(f_h\) and \(f_w\) represent the skeleton video's height and width. The offset \(b_x\) is calculated as the mean of the differences:

\begin{equation}
b_x = \text{mean}(x^r - x^d \cdot a_x)
\end{equation}

Finally, the affine transformation for each point \((x, y)\) is applied following:
\begin{equation}
\begin{pmatrix} x_i' \\ y_i' \end{pmatrix} = \begin{pmatrix} a_x & 0 \\ 0 & a_y \end{pmatrix} \begin{pmatrix} x_i \\ y_i \end{pmatrix} + \begin{pmatrix} b_x \\ b_y \end{pmatrix}
\end{equation}
where \((x_i', y_i')\) is the transformed keypoints.

After using the affine transformation, the position and the scale of the generated skeleton could be similar to the reference image. However, the specific length of neck, arms or legs are not exactly the same with the reference image,as the affine transformation apply a set of parameters for all skeleton points in the generated skeleton array. To solve this problem, we proposed to select an anchor point and fix it, then we separately rescale the neck, arms, hands and legs to the exactly same length with the reference image for better alignment. To be specifically, we select the neck point as the anchor point and then rescale the length of left arm, we translate one point at a time and all the other points linked to this point are carried out the same translation, thus we do not break the relative positions between different skeleton points. After rescaling each frame separately, we transform them into a skeleton video with the same scale and exactly the same length of body parts with the reference image.

\begin{figure}[t]
  \centering
  \includegraphics[width=1.\linewidth]{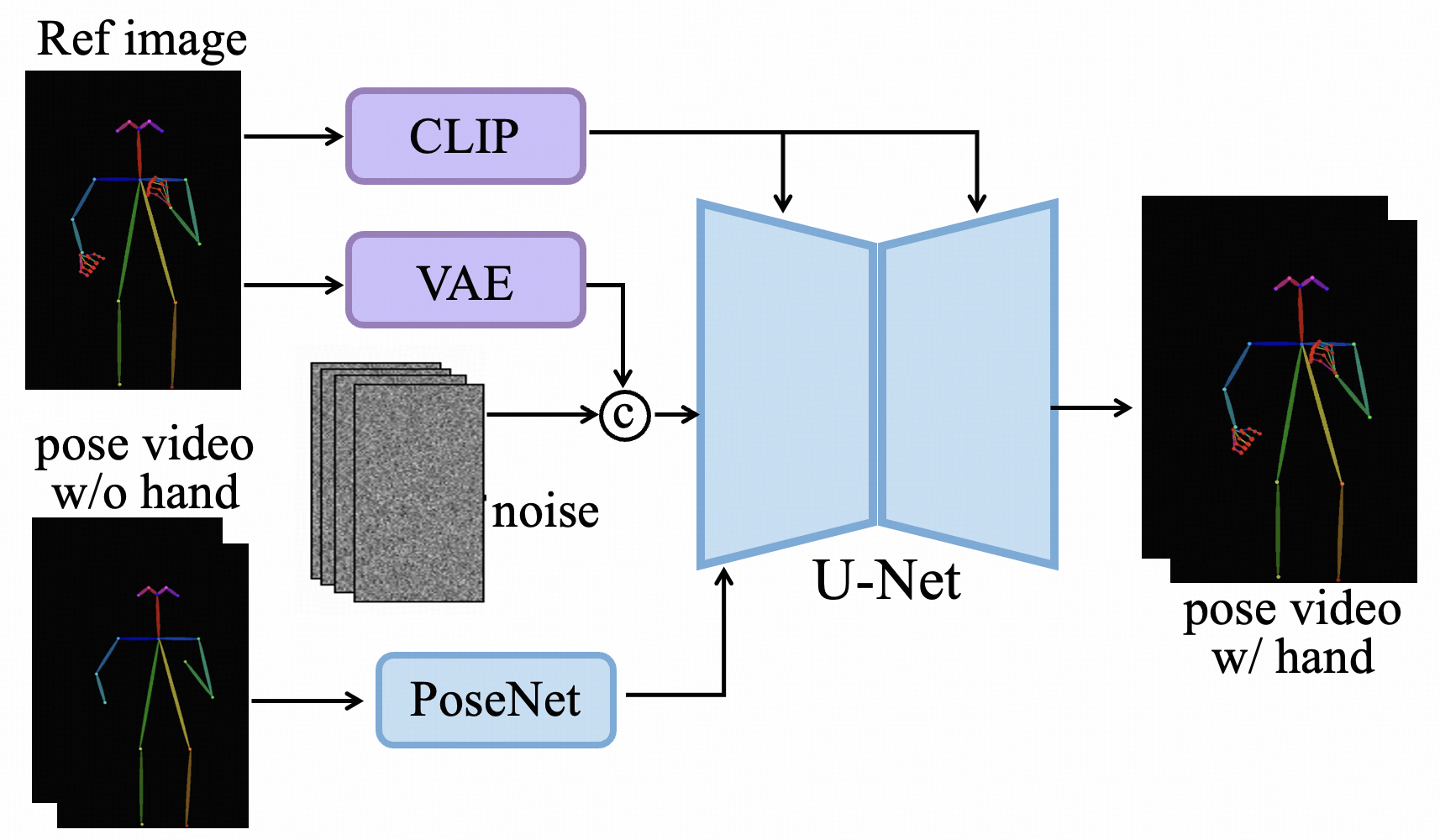}
  \caption{The structure of our proposed skeleton adapter. The CLIP~\cite{radford2021learning} and VAE~\cite{kingma2013auto} are fixed during training, while the PoseNet is trained from scratch, the U-Net~\cite{ronneberger2015u} is fine-tuned from Stable Video Diffusion~\cite{blattmann2023stable}. The reference image is sampled randomly from training pose videos with hands and we use the pose videos without hands for motion guidance. During inference, reference image is detected from given image and the handless pose video is generated by text-to-motion module.}
  \label{skeletonadapter}
\end{figure}

\subsection{Skeleton Adapter}
To further bridge the gap between text-to-motion model and motion-to-video model, we need to compensate the missing information for the text-generated skeleton video. As for the missing face information, we propose to fine-tune the video generation model using the same format skeleton with the T2M-GPT generated ones. We find that only using 5 keypoints including left ear, right ear, left eye, right eye and nose can already get good face in the generated videos, the results are shown in Sec.~\ref{ablationstudysection}. However, the missing hand information cannot be introduced by fine-tuning as different hand motions like clench fist or open palm cannot be simply represented by one keypoint.

To solve this problem, we design a skeleton adapter to generate hand skeletons based on handless skeleton videos, the structure is shown in Fig.~\ref{skeletonadapter}. Our skeleton adapter is inspired by the training processes of image-to-video models. The motivation is to animate a skeleton image with hands (ref image in Fig.~\ref{skeletonadapter}). The animate motion should follow the motion from the handless skeleton videos (pose video w/o hand in Fig.~\ref{skeletonadapter}). To achieve this, we extract two types of pose videos from our training video: those without hands and those with hands. The videos without hands serve as motion guidance for the reference skeleton image, which is randomly extracted from the videos with hands during the training. The primary objective of the skeleton adapter is to recover these pose videos with hands from noise, leveraging both the motion guidance video and a reference skeleton image. During the inference phase, we extract the reference skeleton image from the provided reference human image. The handless pose videos are generated using the text-to-motion module. 

In Fig.~\ref{skeletonadapter}, the primary structure of skeleton adapter is a latent video diffusion model~\cite{rombach2022high} equipped with a U-Net that performs iterative denoising within latent space. We utilize frozen CLIP and VAE to encode the reference skeleton image. The features extracted by CLIP is incorporated via the cross-attention layers in every U-Net block. The features extracted by the pre-trained VAE encoder is replicated across the temporal dimension and then concatenated with the latent video frames along the channel dimension. We use PoseNet to extract features from the handless pose videos for pose guidance, which is constructed with multiple convolutional layers. The extracted pose features are then added element-wise to the features of noised video input for further diffusion process.

\begin{figure*}[t]
  \centering
  \includegraphics[width=1.\linewidth]{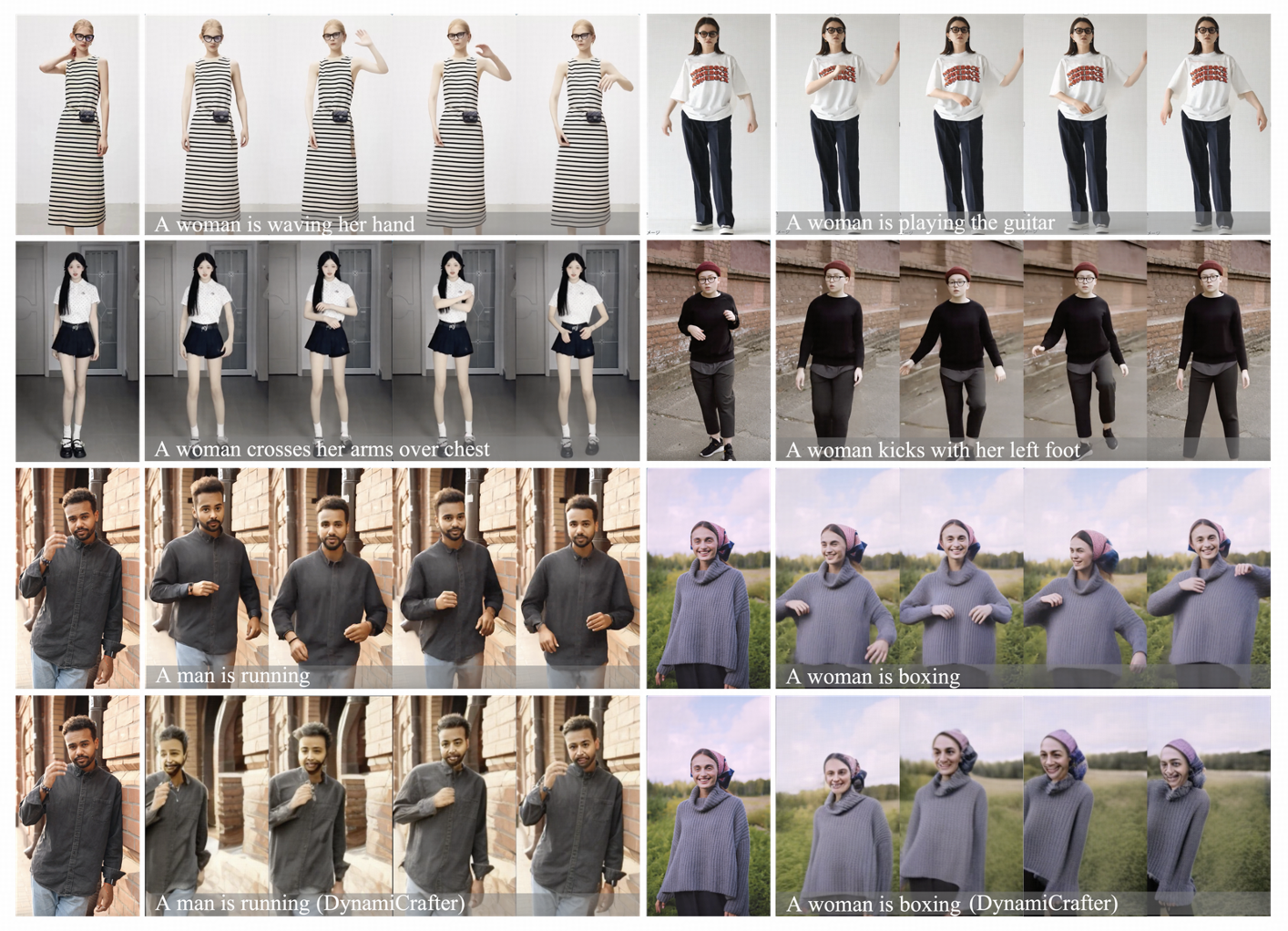}
  \caption{Qualitative results of Fleximo (first three rows) compared to the SOTA text-conditioned image-to-video generation method, DynamiCrafter (last row), in the text-to-human motion video generation task.}
  \label{exp1}
\end{figure*}

\subsection{LLM Planning and Video Refinement}
We propose LLM planning and video refinement, as plug-and-play modules to enhance the video quality. 
LLM planning addresses the generation of long motion videos. Similar to RPG Diffusion~\cite{yang2024mastering}, we use a template and ask an LLM to divide long input motion texts into smaller motion segments. We then generate motion videos corresponding to each segment. We ensure the beginning and ending frames are the reference image, which allows us to concatenate the videos seamlessly into long motion sequences, as illustrated in Fig.~\ref{teaser}.

Video refinement involves generating an initial anchor video using the reference image and text-generated skeleton videos. We then use DWPose~\cite{yang2023effective} to extract skeleton videos from the anchor video. The extracted skeleton video, along with the reference image, is subsequently used to generate the final human motion video. The rationale behind refinement is that the format of the text-generated skeleton videos differs from that of the DWPose-extracted skeleton videos, particularly with respect to the face points. These points are not rescaled using anchor-point-based rescaling method, as they involve rotation that cannot be corrected by simple translation. Since the motion-to-video module is trained using DWPose-extracted skeleton videos, we apply the refinement technique to improve the quality of the generated video.

\section{MotionBench}

Our objective is to generate human motion videos using only reference images and text inputs. 
However, existing datasets and benchmarks like TikTok~\cite{jafarian2021learning} or even VBench~\cite{huang2023vbench} fall short when it comes to assessing how well the generated videos align with a diverse range of specific motions. To address this, we introduce a new benchmark named MotionBench, comprising 20 unique identities performing 20 different motions, resulting in 400 distinct human motion videos. Reference images for the different identities are collected online or generated using Kling AI\footnote{\url{https://klingai.com}}. More details are in the supplementary material.

As existing metrics do not adequately assess how well the generated videos aligns with the motion text, we also propose a new metric named MotionScore. 
Specifically, we input videos from MotionBench into MotionLLM~\cite{chen2024motionllm}, prompting it to "describe the motion in the video," and obtain an LLM-generated response. We then calculate a cosine similarity between this text response and the motion text in the CLIP embedding space. The score reflects the similarity between the LLM’s description and the original motion text, a higher score suggests that the generated motion is more aligned with the motion text.

\textbf{Discussion} We also experimented with other alternatives. For instance, given a prompt such as "The person in the video is running, yes or no?", we tasked the LLM with binary classification to gauge motion accuracy. However, this approach often resulted in biased "yes" responses, yielding artificially high accuracy and diminishing the metric's reliability. Another approach involved asking the LLM to choose between options, such as "Is the person running or playing the guitar?" Unfortunately, this often led to ambiguous responses like "The person is running while waving their hands as if playing the guitar," introducing further confusion. We also explored multi-class classification for various motion categories, but the results from MotionLLM were unsatisfactory. After exploring these alternatives, we determined that the most reliable metric is MotionScore.

\begin{table*}[t]
\begin{center}
\caption{Quantitative comparison with other SOTA methods. We use our proposed MotionScore to evaluate the motion following of different video generation methods.}
\label{quantitative}
\setlength{\tabcolsep}{2mm}
\begin{tabular}{c|ccccc|c|c}
\Xhline{1pt}
Methods & PSNR$\uparrow$&  SSIM$\uparrow$ & LPIPS$\downarrow$  & DreamSim$\uparrow$ &  FID$\downarrow$
            & FVD $\downarrow$ & MotionScore $\uparrow$ \\ \Xhline{1pt}
I2VGen-XL~\cite{zhang2023i2vgen}    & 7.931 &  0.3684       & 0.559  & 0.537   &  220.221    & 1905.26 & 0.6806  \\
VideoCrafter~\cite{chen2023videocrafter1} & 6.727 & 0.5190 &  0.608   &  0.211   &  149.534  &  1536.00 & 0.6866  \\

DynamiCrafter~\cite{xing2023dynamicrafter} & 9.607 & 0.6800 & 0.407   &  0.699   &  99.644   & 1462.42 & 0.6868 \\
Fleximo &  \textbf{16.647} & \textbf{0.7148} & \textbf{0.284} & \textbf{0.879} &  \textbf{76.181}    &  \textbf{1360.12}  &  \textbf{0.6990}  \\ 
\Xhline{1pt}
\end{tabular}
\vspace{-2mm}
\end{center}
\end{table*}

\section{Experiments}

\subsection{Implementation Details}
We train our model on HumanVid~\cite{wang2024humanvid} dataset which contains around 8K high-quality vertical human motion videos collected from the internet. The weights of U-Net and PoseNet are initialized from the MimicMotion. The learning rate is $10^{-5}$ with no linear warmup. We train the model on 8 NVIDIA H100 GPUs (80G) with the resolution of 1024 (height) $\times$ 576 (width) for 50K steps, the frame length is 14, and the batch size is 1 for each device. The parameters of U-Net and PoseNet are all fine-tuned.

\subsection{Comparison with Other SOTA Methods}
In this section, we evaluate Fleximo both qualitatively and quantitatively. We compare it with state-of-the-art (SOTA) text-conditioned image-to-video methods, including I2VGen-XL~\cite{zhang2023i2vgen}, VideoCrafter~\cite{chen2023videocrafter1}, and DynamiCrafter~\cite{xing2023dynamicrafter}, which can generate motion videos based on a reference image and motion text. We show the qualitative results in Fig.~\ref{exp1}. The reference images are sourced from the internet and consist entirely of out-of-domain images that were not seen during model training. We present six different IDs performing six distinct motions. We also show the results of the SOTA method, DynamiCrafter, in the last row for comparison. The results demonstrate that Fleximo not only has better video quality, face, and hand details compared to DynamiCrafter but also generates realistic and large movements, as opposed to the small movements of DynamiCrafter. Due to space limitation, more results of Fleximo and other SOTA methods are provided in the Supp. material.

To quantitatively evaluate Fleximo, we carry out experiments on MotionBench. We generate 400 videos using each method and compute average metrics on the 400 videos. We employ several metrics, such as PSNR~\cite{hore2010image}, SSIM~\cite{hore2010image}, LPIPS~\cite{zhang2018unreasonable}, DreamSim~\cite{fu2023dreamsim}, FID~\cite{heusel2017gans}, FVD~\cite{unterthiner2019fvd} and MotionScore. PSNR, SSIM, and LPIPS are metrics used to assess image similarity. PSNR measures image quality, SSIM evaluates luminance, contrast, and structural information, while LPIPS assesses similarity by analyzing feature representations. DreamSim bridges the gap between "low-level" metrics (e.g., LPIPS, PSNR, SSIM) and "high-level" metrics (e.g., CLIP), providing a metric to measure both semantic and perceptual similarity. FID and FVD assess the overall quality of generated videos. For each generated video, we randomly select 10 frames and calculate the metrics against the reference image for each frame, then average the scores. The results shown in Tab.~\ref{quantitative} demonstrate that Fleximo generates frames that closely resemble the reference image in quality (PSNR, SSIM, LPIPS) and high-level semantics (DreamSim). Additionally, Fleximo produces videos of superior quality (FID, FVD). The MotionScore in Table~\ref{quantitative} demonstrates that Fleximo generates videos more closely aligned with the input motion texts.

\subsection{Generated Pose Videos from Skeleton Adapter}
\begin{figure}[t]
  \centering
  \includegraphics[width=1.\linewidth]{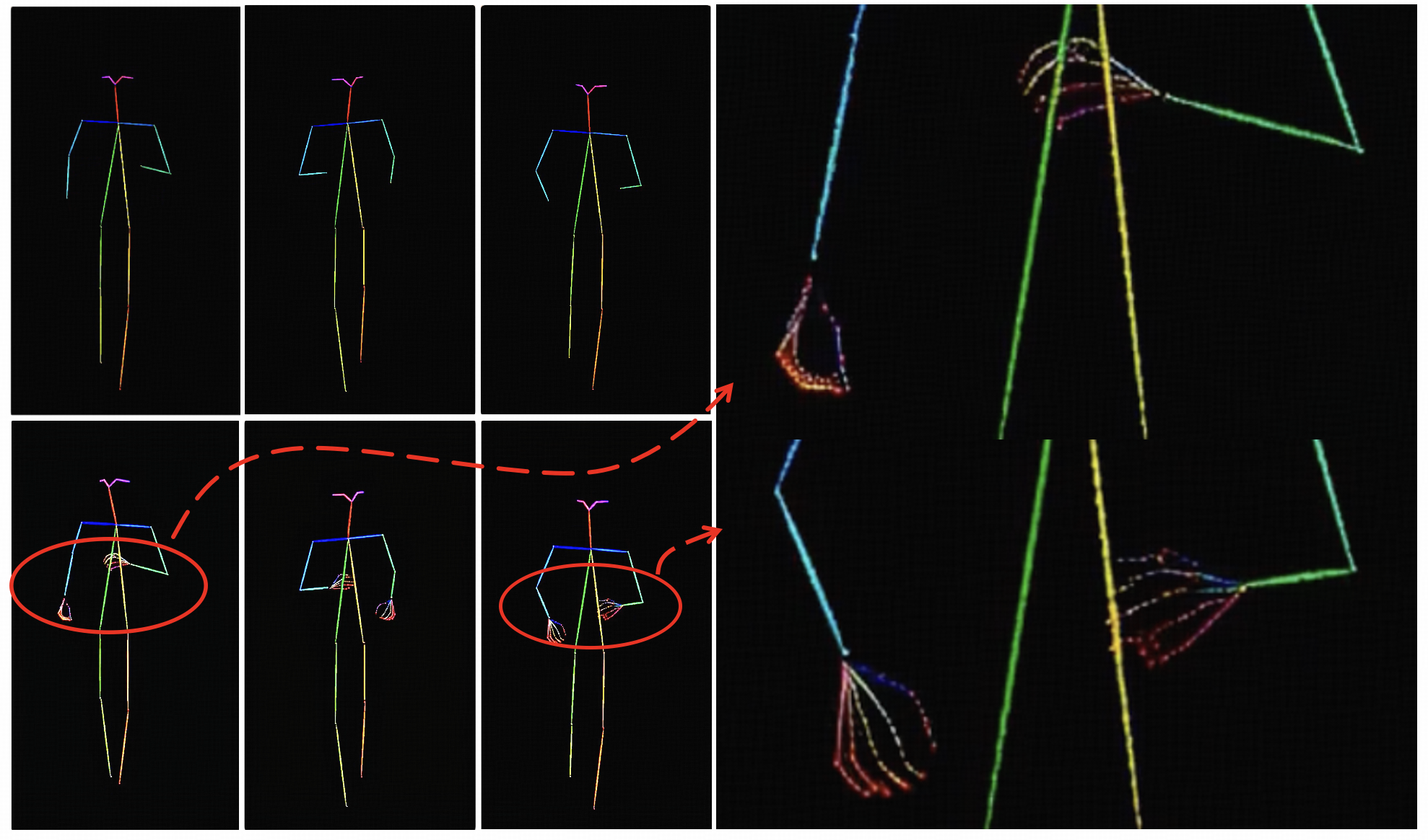}
  \caption{The generated pose video (the second row) of skeleton adapter given the handless pose video (the first row). Skeleton adapter can maintain the motion in the handless pose video while completing detailed and realistic hand information. \textbf{Please zoom in for better visualization.}}
  \label{pose}
  \vspace{-1mm}
\end{figure}

The skeleton adapter is very important for the success of Fleximo as it bridges the gap between text-to-motion module and the motion-to-video module. We show generated pose videos in Fig.~\ref{pose} to illustrate the effectiveness of skeleton adapter. From the results, we can see that skeleton adapter can complete detalied and realistic hand information while maintaining the motion provided in the handless pose videos.

\begin{figure}[t]
  \centering
  \includegraphics[width=1.\linewidth]{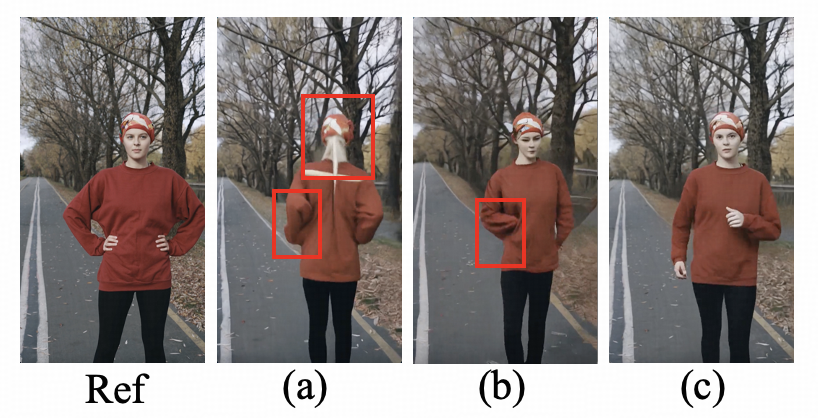}
    \caption{The ablation study of Fleximo. Reference image is shown in the left and the motion text is "a woman is running". (a) shows the results of our baseline, which simply combine T2M-GPT and MimicMotion leading to bad results without either face or hand information. (b) displays the results when we finetune MimicMotion using the same skeleton format of T2M-GPT, there is still no hand information. (c) illustrates the results achieved using skeleton adapter, which completes the hand information and leads to higher-quality motion videos.}
    \vspace{-2mm}
    \label{ablation1}
\end{figure}

\begin{table}[t]
\begin{center}
\caption{User study results of different methods regards video quality, identity preservation and motion-text alignment.}
\label{user}
\setlength{\tabcolsep}{0.6mm}
\begin{tabular}{c|ccc}
\Xhline{1pt}
Methods & Video Quality&  Identity Pres. & Motion Align.\\ \Xhline{1pt}
I2VGen-XL   & 1.21 &  1.46      & 1.35  \\
VideoCrafter & 1.52 & 1.71 &  1.39  \\
 
DynamiCrafter & 1.89 & 2.15 & 1.63   \\
Fleximo &  \textbf{3.77} & \textbf{4.17} & \textbf{3.93} \\ 
\Xhline{1pt}
\end{tabular}
\vspace{-7mm}
\end{center}

\end{table}

\subsection{User Study}
To estimate the quality of generated videos from human perspectives, we conduct a blind user study with 8 participants. Specifically, we utilize the methods in Tab.~\ref{quantitative} to generate 800 videos, each method with 200 videos. We ask each participant to score each generated video from three aspects: video quality, identity preservation and motion-text alignment. The score ranges from 1 to 5, with 5 represents the best quality. The results in Tab.~\ref{user} are averaged on 200 videos, where Fleximo noticeably outperforms other methods in all aspects, demonstrating its superiority and effectiveness. More details are in the Supp. material.

\subsection{Ablation Study}
\label{ablationstudysection}
To study the effectiveness of the proposed modules in Fleximo, we carry out thorough ablation studies in Fig.~\ref{ablation1}. From the results in Fig.~\ref{ablation1} (a), we can see that directly combining SOTA motion-guided video generation methods like MimicMotion with text-to-motion method T2M-GPT cannot generate reasonable human motion video. The reasons lie in that motion-guided video generation method need the full-body skeleton keypoints as the guidance, which include 68 face points, 42 hand points and 18 body points. While T2M-GPT can only generate the skeleton video containing the 18 body points. The information loss of face and hand keypoints degrades the motion video generation. To improve the performance, we further finetune MimicMotion with 18-keypoint skeleton videos to align it with T2M-GPT. The results are shown in Fig.~\ref{ablation1} (b). After finetuning, MimicMotion can generate better results regarding face as the guidance format of skeleton video is now aligned with the output of T2M-GPT.  However, the hands of the generated videos are very blur or in a mess as the guided skeleton video only has one point to represent the hands region. We speculate that the reason lies in that face is a whole part without finer details so it can be represented by only 5 points like in T2M-GPT format. However, the hand information is much more fine-grained including the fingers. Thus, it is impossible to generate hand details using only one point (T2M-GPT format) as guidance. Combined with skeleton adapter, the results in Fig.~\ref{ablation1} (c) contain both face and hands details, which are more realistic and with better visual quality.

\begin{figure}[t]
  \centering
  \includegraphics[width=.9\linewidth]{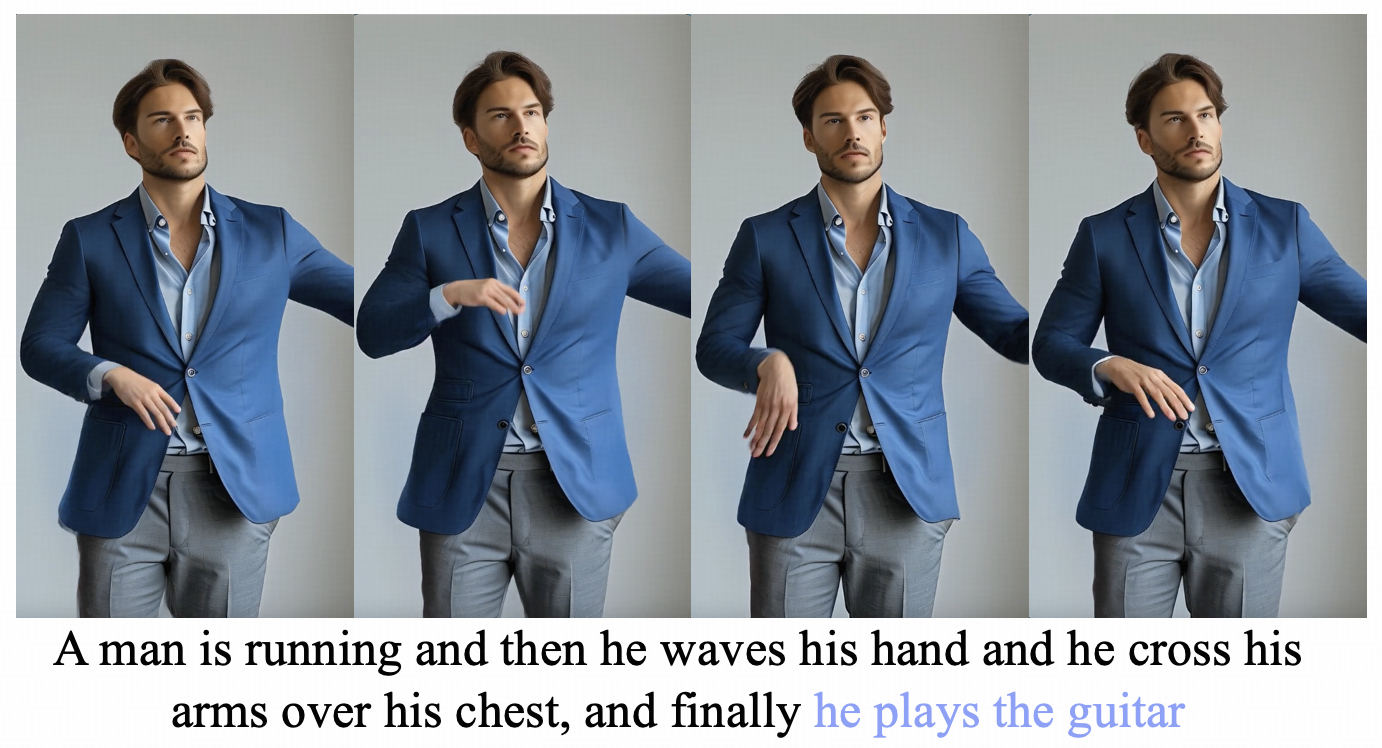}
    \caption{The generated video could only respond to a small motion piece (marked by blue) and neglect other inputs (marked by black) without using LLM planning.}
            \vspace{-2mm}
    \label{llmplan}
\end{figure}

We also validate the effectiveness of our introduced LLM planing. Shown in Fig.~\ref{llmplan}, without LLM Planning, Fleximo cannot understand several motions in one motion text, and it only generates a short motion sequence corresponding to the end of the motion text. Using LLM planning, the long motion texts are separated into several consecutive short motions, and we can generate long motion videos shown in Fig.~\ref{teaser}.

\begin{figure}[t]
  \centering
  \includegraphics[width=1.\linewidth]{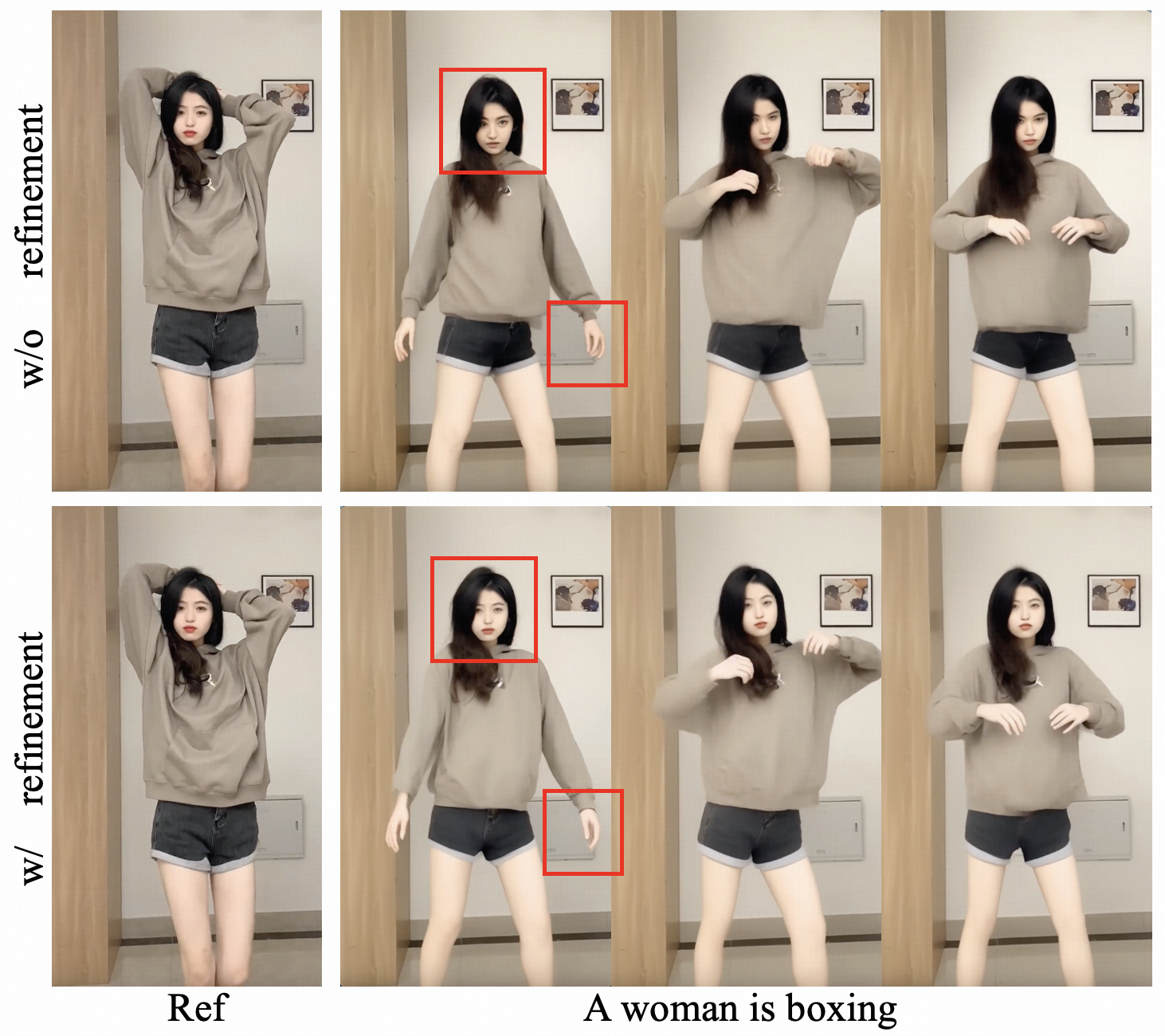}
    \caption{Generated videos with and without refinement. With refinement, the generated videos have better quality, particularly in terms of facial identity preservation and hand generation.}
    \vspace{-2mm}
    \label{refinement}
\end{figure}

We examine the effectiveness of our refinement in Fig.~\ref{refinement}. Without refinement, which means we directly use the text-generated skeleton videos for guidance, the preservation of facial identity is limited. The reason lies in that the text generated skeleton videos may not align well with the DWPose extracted skeleton videos. Using refinement, the generated videos have better facial identity consistency. We also observe that using anchor video results in better hand generation, as marked by the red rectangle in Fig.~\ref{refinement}.

\begin{figure}[t]
  \centering
  \includegraphics[width=1.\linewidth]{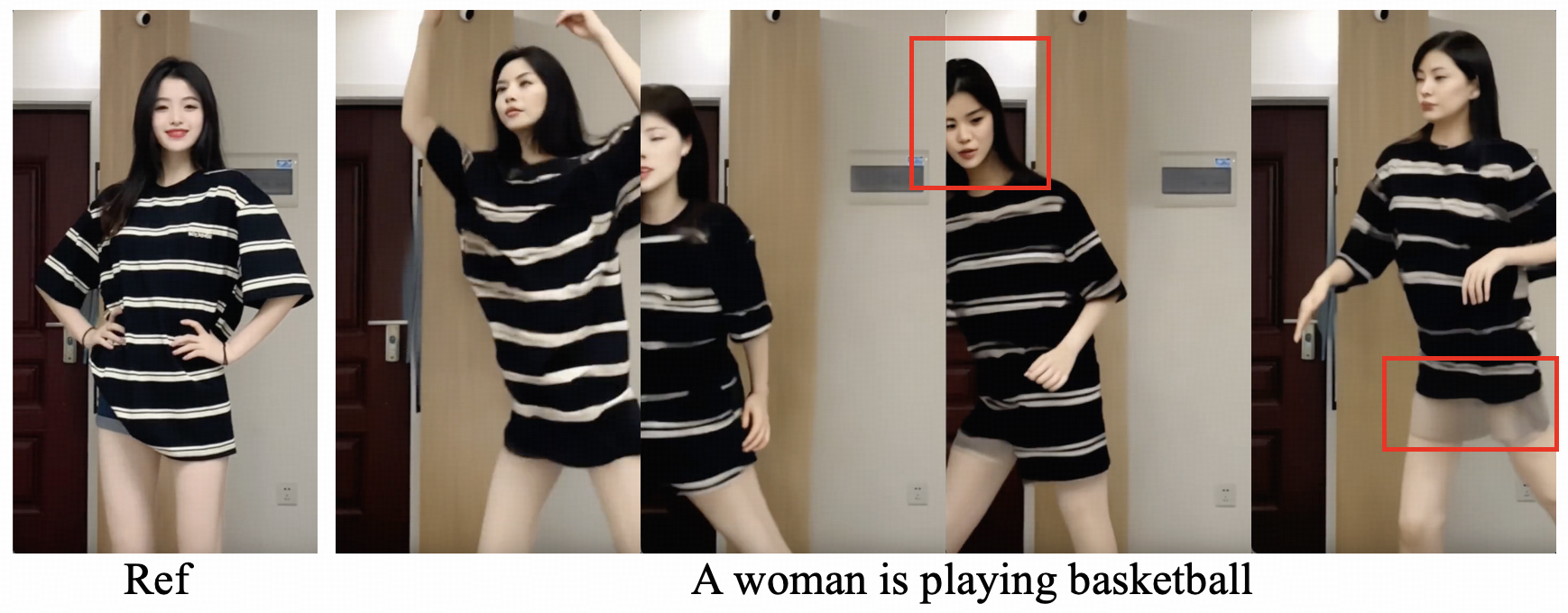}
    \caption{Failure case of Fleximo. When the motion involves large movements, the skeleton scaling becomes inconsistent, leading to scale jitter and poor identity preservation.}
        \vspace{-1mm}
    \label{failure}

\end{figure}

\subsection{Failure Cases and Limitations}

While Fleximo can generate high-quality human motion videos from reference images and motion texts, it does have limitations. For instance, it struggles with generating large-scale movements, which is shown in Fig.~\ref{failure}. Given a text prompt like "A woman is playing basketball", the text-to-motion module produces a pose video where the human figure undergoes significant positional shifts, including side-facing movements. This variability makes it challenging for the motion-guided video generation module to maintain facial identity and body consistency, as it is highly sensitive to the position and scale of the guided skeleton. Additionally, the text-to-motion model focuses solely on generating body movements without interactions with objects, which reduces the realism of the generated videos. For example, it may depict someone "playing the guitar" with only arm movements, without including an actual guitar.

\section{Conclusion}
In this paper, we introduce a new task: text-to-human motion video generation, which is more flexible and user-friendly than pose video guidance. To address this task, we propose \textbf{Fleximo}, a novel framework that leverages the large-data pretrained text-to-3D motion model, incorporating anchor point-based rescaling, skeleton adaptation, anchor video refinement, and LLM planning. Additionally, we introduce MotionBench, a benchmark with diverse human identities and motions, along with the MotionScore metric to assess motion-text alignments. Fleximo demonstrates superior performance in generating high-fidelity human motion videos, outperforming several leading text-conditioned image-to-video models both qualitatively and quantitatively.

{
    \small
    \bibliographystyle{ieeenat_fullname}
    \bibliography{main.bbl}
}


\clearpage
\setcounter{page}{1}


\twocolumn[{%
\renewcommand\twocolumn[1][]{#1}%
\maketitlesupplementary
\begin{center}
\includegraphics[width=0.95\linewidth]{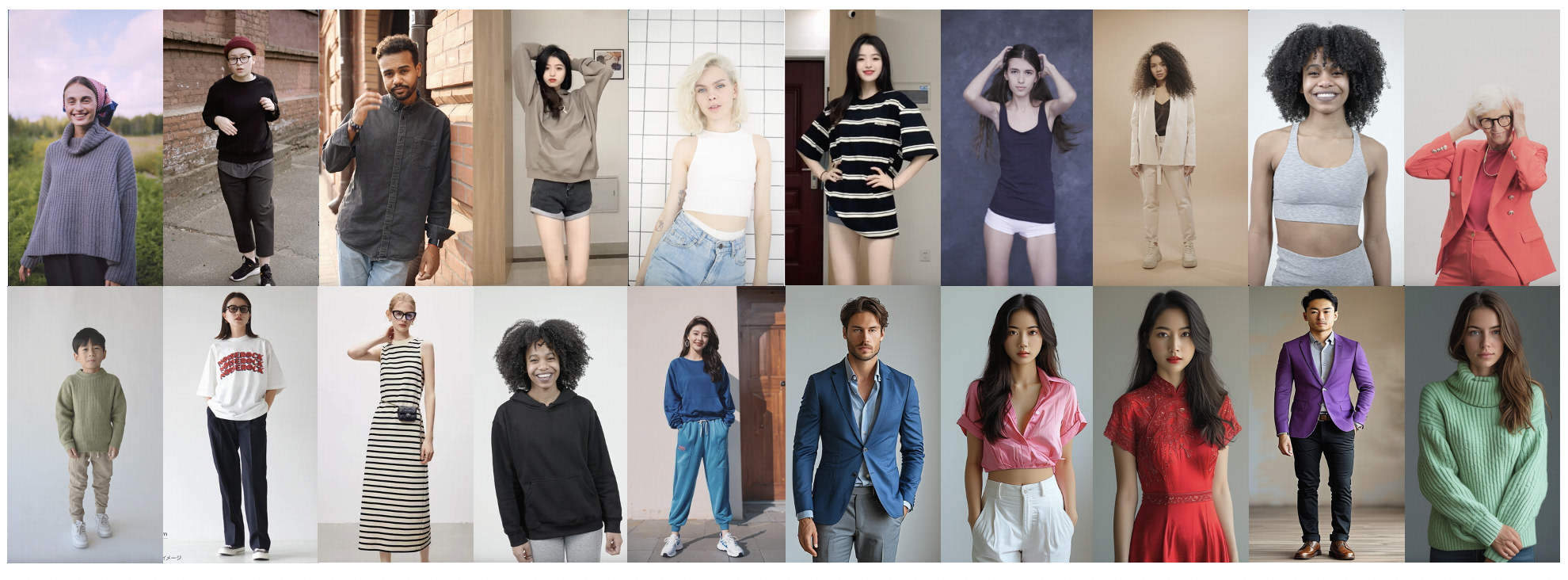}
\captionof{figure}{The identities in our MotionBench. The last 5 identities are generated by Kling AI, and all other identities are sourced online.}
\label{s1}
\end{center}
}]


\section{MotionBench}

We present 20 distinct identities in MotionBench, as shown in Fig.~\ref{s1}. These identities represent a diverse spectrum of individuals, encompassing a range of genders, ages, and ethnic backgrounds, ensuring that our dataset is comprehensive and inclusive. The identities are carefully curated to reflect real-world diversity and to be representative of the varied human forms and appearances that researchers might encounter in practical applications. The images in MotionBench are sourced either from publicly available online datasets or are generated using Kling AI (for the last five identities), ensuring high variability and rich visual content.

The images themselves feature a wide array of settings: some include backgrounds, while others are background-free, providing flexibility for different types of motion analysis. Additionally, the images showcase both full-body and partial-body shots, further increasing the diversity of the data and enabling researchers to focus on specific body parts or motions. These variations in body pose, camera angle, and scene context allow MotionBench to support a broad range of tasks, including but not limited to pose estimation, motion recognition, and action classification.

\begin{table*}[t]
\centering
\setlength{\tabcolsep}{3mm}
\begin{tabular}{c|l|c|l}
\Xhline{1pt}
Number & Motion Text & Number & Motion Text \\ \Xhline{1pt}
1  & A person is waving hand. & 2  & A person is raising leg. \\ \hline
3  & A person is bending knees. & 4  & A person is playing the violin. \\ \hline
5  & A person is crossing arms over chest. & 6  & A person is talking. \\ \hline
7  & A person is kicking with left foot. & 8  & A person is raising hand to chest. \\ \hline
9  & A person is brushing hair. & 10 & A person is lifting right foot. \\ \hline
11 & A person is moving hand in circle. & 12 & A person is warming up. \\ \hline
13 & A person is running. & 14 & A person is wiping. \\ \hline
15 & A person is raising arm. & 16 & A person is walking. \\ \hline
17 & A person is playing golf. & 18 & A person is jumping. \\ \hline
19 & A person is boxing. & 20 & A person is playing the guitar. \\ \hline
\Xhline{1pt}
\end{tabular}
\caption{List of 20 different motions in MotionBench. For simplicity, we replace "man/woman" with "person" in the motion text.}
\label{20motions}
\end{table*}

For each identity, we generate 20 different motion sequences, capturing a wide range of human actions. These motions cover both common everyday activities and more dynamic movements, ensuring that the dataset is useful for a variety of motion analysis tasks. The 20 motion types are listed in Tab.~\ref{20motions} and include actions such as: waving the hand, raising a leg, bending the knees, playing musical instruments, and performing sports-related actions. These actions span a variety of functional movements, such as lifting a foot or crossing arms, as well as interactive behaviors, such as talking and wiping. Notably, some of the motions represent more dynamic, complex activities, such as boxing, jumping, and playing golf, which can be used to study more complex motion patterns and transitions. The inclusion of these motions ensures that MotionBench serves as a rich, diverse resource for testing and developing algorithms related to human motion understanding.

The wide range of actions and identities in MotionBench makes it an invaluable resource for evaluating motion-related models, enabling researchers to assess their algorithms in more realistic and varied contexts. Furthermore, the variety of motion types—ranging from simple gestures to more complex actions—ensures that MotionBench can support a variety of research objectives in computer vision and machine learning, from basic gesture recognition to more advanced action segmentation and classification tasks.

\section{Results of Fleximo and Other Methods}

We provide more results of Fleximo and other methods in Fig.~\ref{s3lowr}, Fig.~\ref{s4} and Fig.~\ref{s7}. We also provide the original videos in the zip file of the supplementary material. We show the same person performing 10 different motions in Fig.~\ref{s3lowr}, which illustrates that Fleximo can generate diverse motions. To compare with other methods, we show the results generated by Fleximo, I2VGen-XL~\cite{zhang2023i2vgen}, VideoCrafter~\cite{chen2023videocrafter1}, and DynamiCrafter~\cite{xing2023dynamicrafter} in Fig.~\ref{s4} and Fig.~\ref{s7}. The results demonstrate that VideoCrafter only generates small motion or no motion at all, while I2VGen-XL introduces more artifacts when generating the motion. The human motion video generated by Fleximo accurately follows the motion described in the motion text while preserving the identity in the reference image. Additionally, the hand and face details of Fleximo are superior to all other methods.

\begin{figure*}[t]
  \centering
  \includegraphics[width=1.\linewidth]{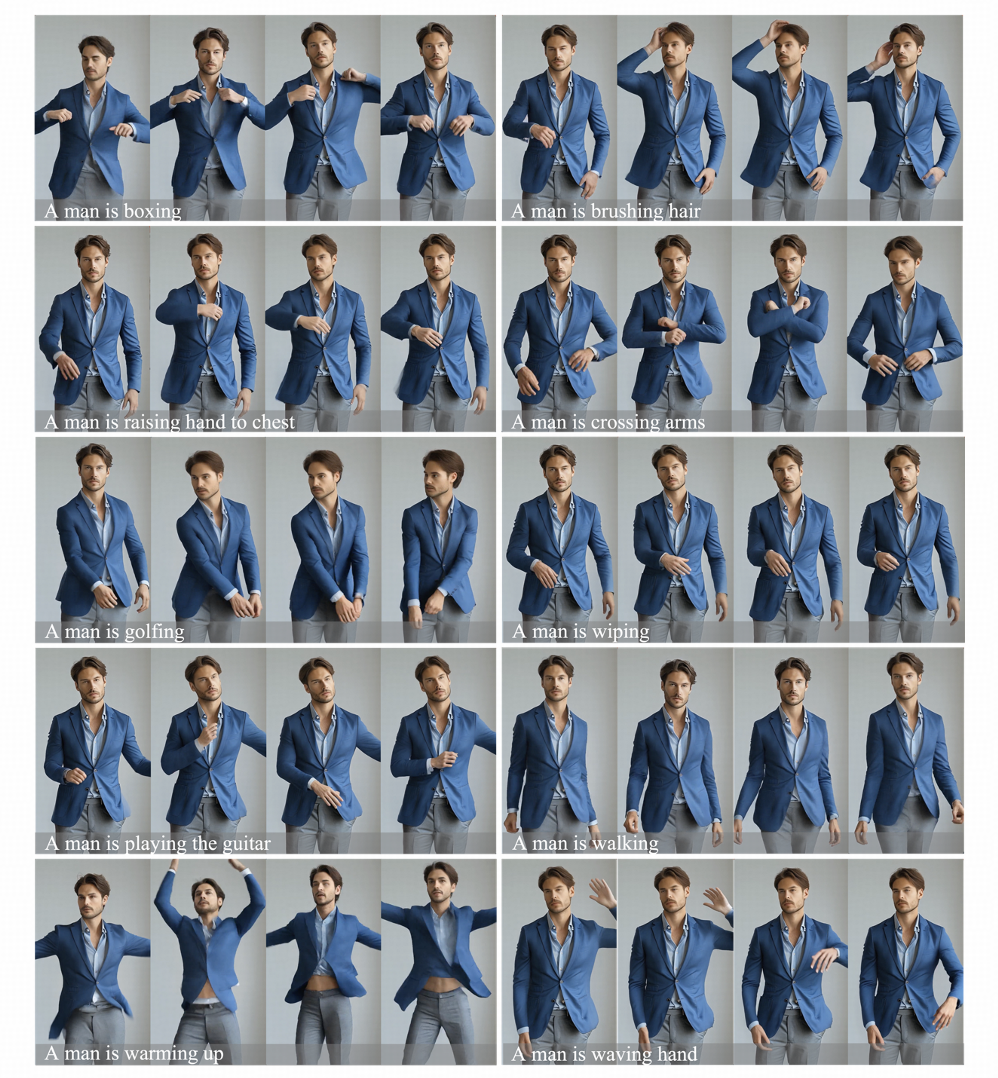}
    \caption{Different motions generated by Fleximo. We show a same identity performing 10 different motions to illustrate the motion diversity of Fleximo.}
    \label{s3lowr}
\end{figure*}

\begin{figure*}[t]
  \centering
  \includegraphics[width=.75\linewidth]{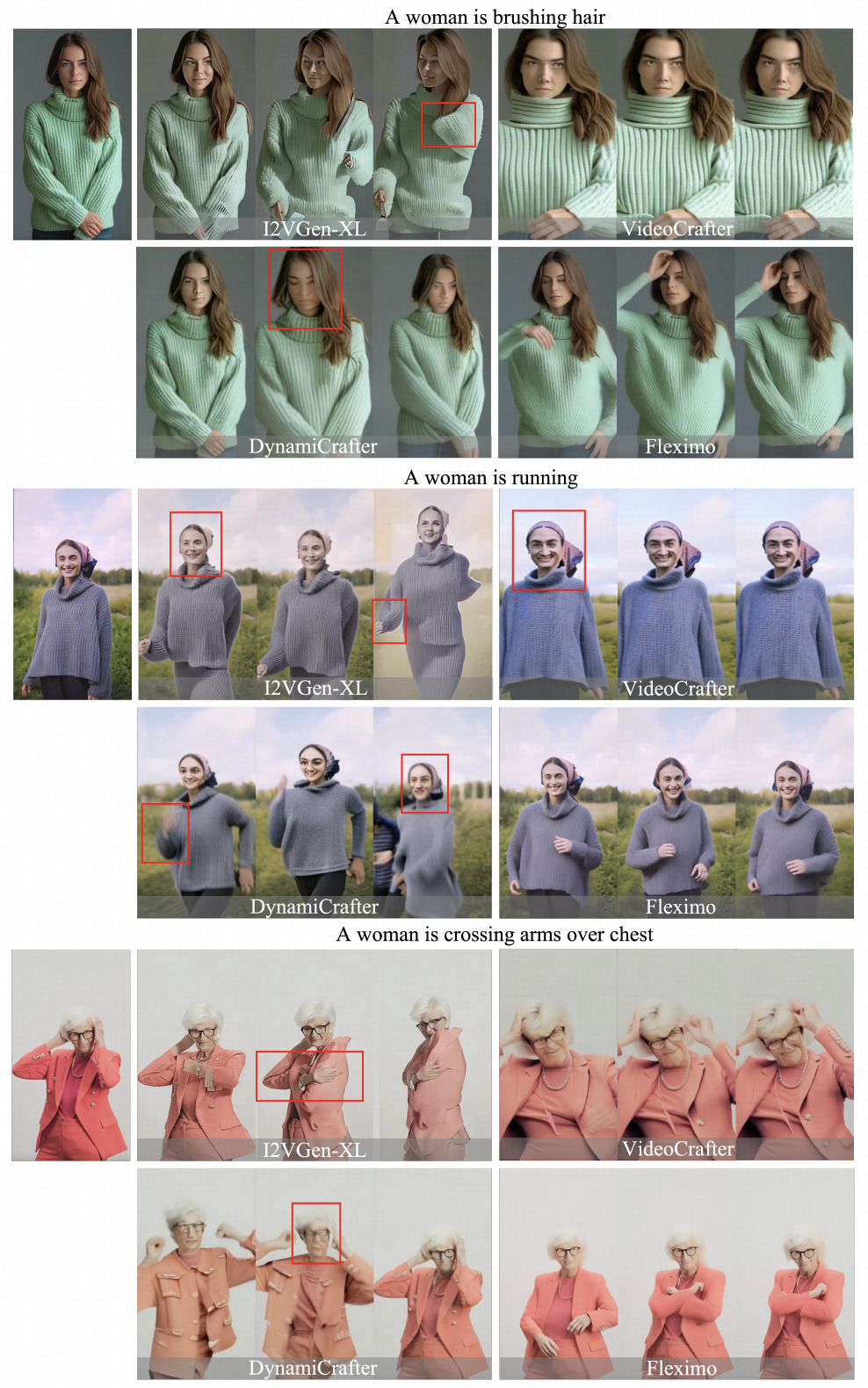}
    \caption{Qualitative comparison with other methods. The reference image is shown on the left. We use various methods to generate human motion videos based on the motion text provided above. Among them, Fleximo produces the highest-quality motion videos.}
    \label{s4}
\end{figure*}

\begin{figure*}[t]
  \centering
  \includegraphics[width=.75\linewidth]{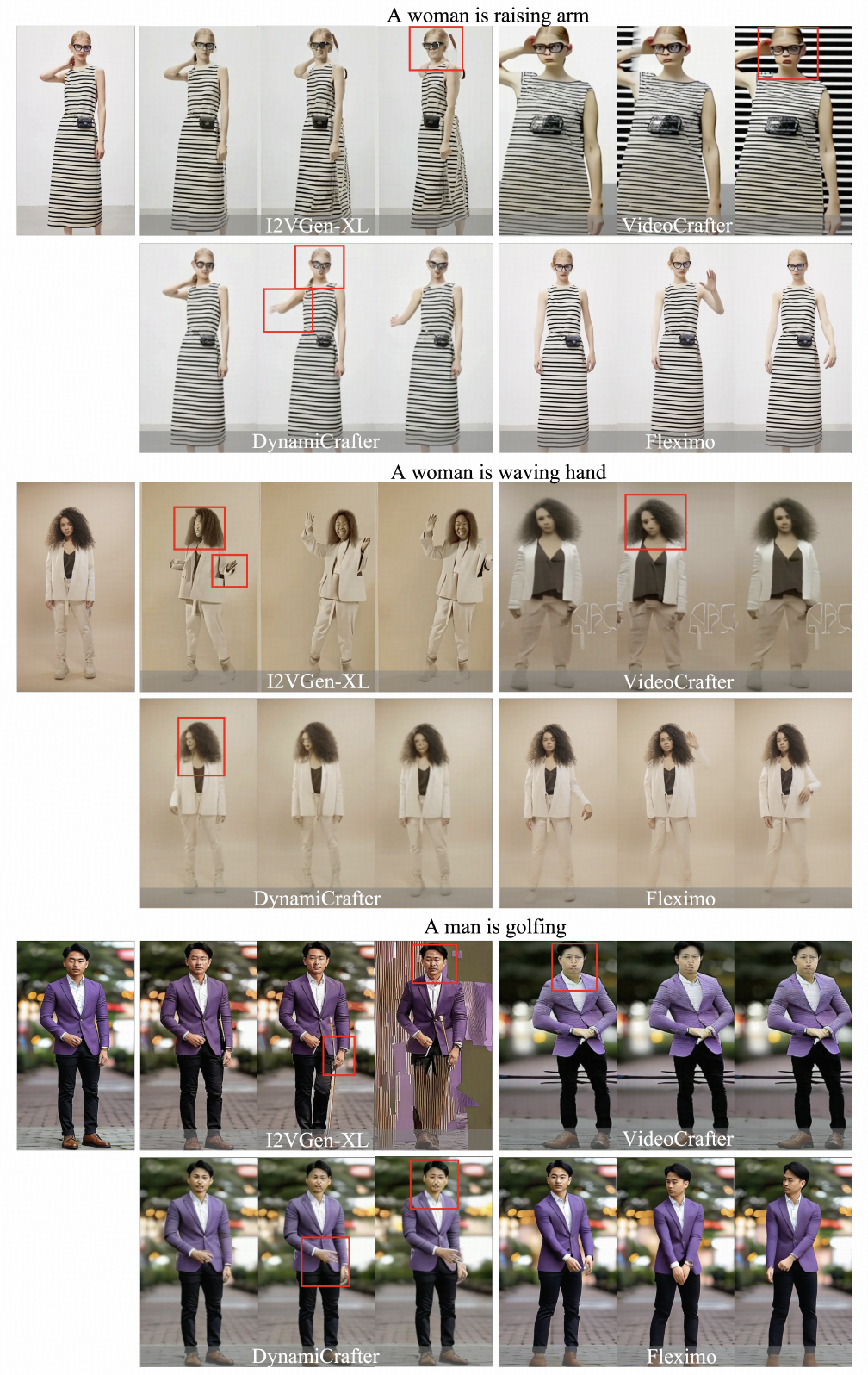}
    \caption{Qualitative comparison with other methods. The reference image is shown on the left. We use various methods to generate human motion videos based on the motion text provided above. Among them, Fleximo produces the highest-quality motion videos.}
    \label{s7}
\end{figure*}

\section{Details of User Study}

We generate 800 videos using four different video generation methods: I2VGen-XL, VideoCrafter, DynamiCrafter, and Fleximo. Eight participants evaluate each video across three criteria: video quality, identity preservation, and motion-text alignment. Each participant scores 100 videos, with four participants being familiar with the video generation field and the other four not having prior expertise in this area, to minimize potential bias. We evaluate the methods from three aspects: video quality, identity preservation, and motion-text alignment.

\begin{itemize}
    \item \textbf{Video quality} refers to the overall quality of the generated video and how closely it resembles real-world videos.
    \item \textbf{Identity preservation} measures how well the generated video maintains the identity from the reference image.
    \item \textbf{Motion-text alignment} assesses how accurately the generated motion follows the input motion text.
\end{itemize}

In Tab. 2 of the original paper, Fleximo achieves the best performance under all three aspects. DynamiCrafter is the second best method. I2VGen-XL is less suitable for human motion video generation as it is basically designed for scene-based video generation. Also, in order to suit its fixed resolution, we resize and pad the reference images, which might degrade the performance of I2VGen-XL to some extent.
We also observe a positive correlation between the three evaluation criteria in Tab. 2 of the original paper: methods that achieve better video quality tend to also perform better in identity preservation and motion-text alignment. Notably, our method shows the most improvement in motion-text alignment, outperforming DynamiCrafter by 2.3 points. This demonstrates that our method generates more coherent and reasonable motions compared to other state-of-the-art (SOTA) text-to-video and image-to-video methods.

\begin{figure*}[t]
  \centering
  \includegraphics[width=1.\linewidth]{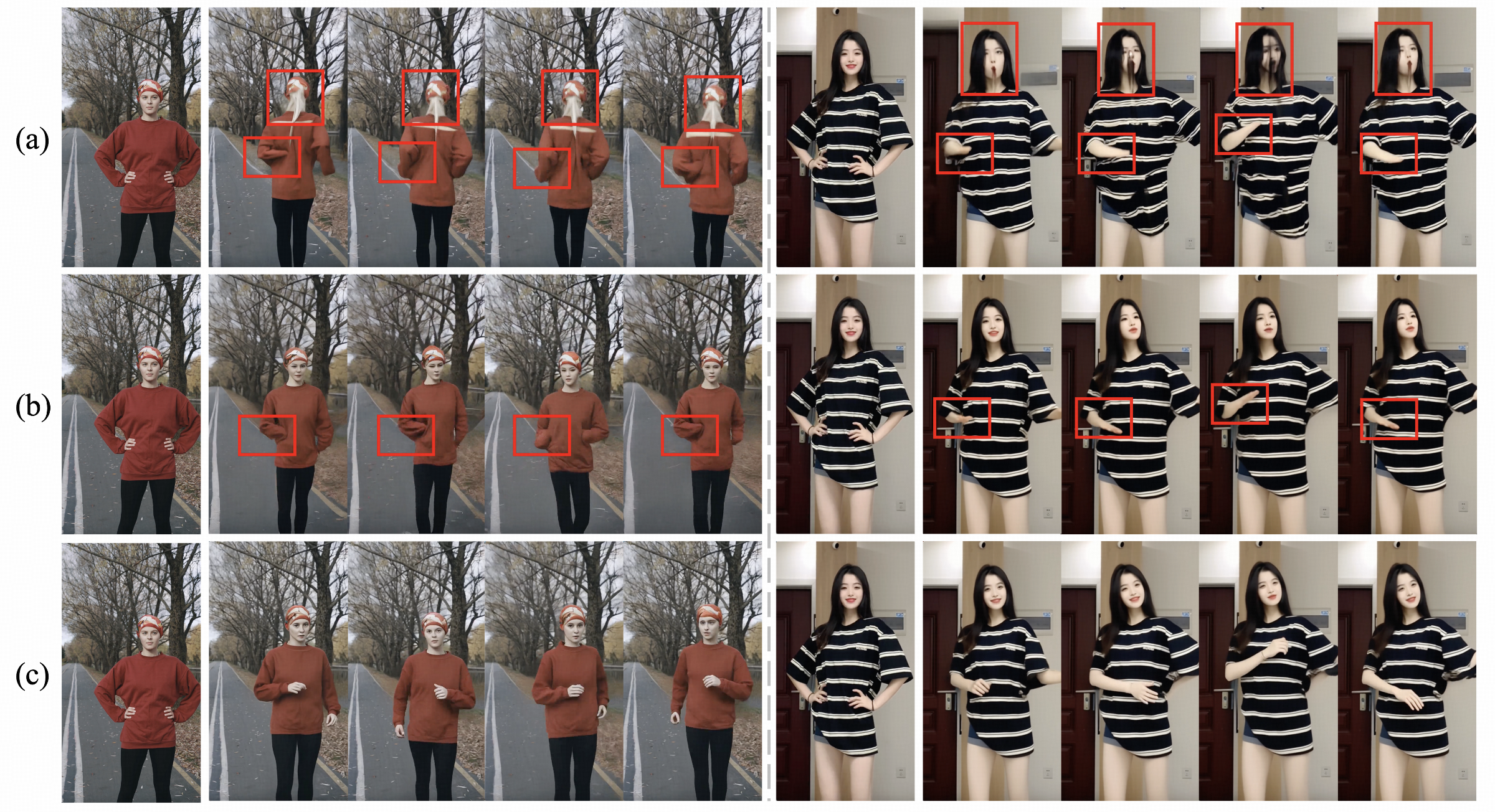}
    \caption{More results of the ablation study. The reference image is presented in the left, and the motion texts are "a woman is running" and "a woman is playing the guitar" (a) shows the results of the baseline, which simply combines T2M-GPT and MimicMotion, leading to poor results without either face or hand information. (b) displays the results when we fine-tune MimicMotion using the same skeleton format as T2M-GPT. The face can be generated well, but there is still no hand information. (c) illustrates the results achieved by Fleximo, which generates both detailed face and hands, resulting in higher-quality motion videos.}
    \label{ablation1}
\end{figure*}

\section{More Results of Ablation Study}

We provide more results of ablation study in Fig.~\ref{ablation1}. As illustrated in Fig.~\ref{ablation1} (a), directly combining the motion-guided video generation method MimicMotion~\cite{zhang2024mimicmotion} with the text-to-motion approach T2M-GPT~\cite{zhang2023generating} does not yield realistic human motion videos. This issue arises because motion-guided methods, such as MimicMotion, require full-body skeleton keypoints—including 68 facial points, 42 hand points, and 18 body points—while T2M-GPT only generates skeleton videos with the 18 body points. The absence of facial and hand keypoints leads to poor motion generation quality. To resolve this, we fine-tune MimicMotion using 18-keypoint skeleton videos to match T2M-GPT’s output, as shown in Fig.~\ref{ablation1} (b). After fine-tuning, MimicMotion produces improved results, especially for the face, as the skeleton format is now consistent with T2M-GPT. However, the hands in the generated videos remain blurry or disorganized because the skeleton only represents the hand region with a single point. We hypothesize that, unlike the face, which can be sufficiently represented by a few points (as done in T2M-GPT), the hands including the fingers cannot be accurately captured with just one point. By introducing our proposed skeleton adapter, as shown in Fig.~\ref{ablation1} (c), we can better capture both facial and hand details, resulting in more realistic and visually appealing human motion videos.

\end{document}